\documentclass[a4paper,fleqn]{cas-dc}

\usepackage[authoryear,longnamesfirst]{natbib}
\usepackage{color}

\def\tsc#1{\csdef{#1}{\textsc{\lowercase{#1}}\xspace}}
\tsc{WGM}
\tsc{QE}
\tsc{EP}
\tsc{PMS}
\tsc{BEC}
\tsc{DE}
\begin{document}
\let\WriteBookmarks\relax
\def\floatpagepagefraction{1}
\def\textpagefraction{.001}
\let\printorcid\relax


\shorttitle{Invisible Gas Detection: An RGB-Thermal Cross Attention Network and A New Benchmark}

\shortauthors{Wang and Lin et~al.}
\title [mode = title]{Invisible Gas Detection: An RGB-Thermal Cross Attention Network and A New Benchmark}
                   
% \tnotemark[1,2]

% \tnotetext[1]{This document is the results of the research
   % project funded by the National Science Foundation.}

% \tnotetext[2]{The second title footnote which is a longer text matter
   % to fill through the whole text width and overflow into
   % another line in the footnotes area of the first page.}

\author[1,3]{\textcolor{black}{Jue Wang}}

% \cormark[1]

\fnmark[1]

% \ead{cvr_1@tug.org.in}

% \ead[url]{www.cvr.cc, cvr@sayahna.org}

\credit{Methodology, Data Collection, Experiment, Writing}

\affiliation[1]{organization={Southern University of Science and Technology},
    % addressline={Radarweg 29},
    city={Shenzhen},
    % citysep={}, % Uncomment if no comma needed between city and postcode
    postcode={518055}, 
    state={Guangdong},
    country={China}}

\author[2]{\textcolor{black}{Yuxiang Lin}}

\fnmark[1]

% \ead{cvr3@sayahna.org}
% \ead[URL]{www.sayahna.org}
\credit{Methodology, Data Collection, Writing}
\affiliation[2]{organization={Shenzhen Technology University},
    city={Shenzhen},
    % citysep={}, % Uncomment if no comma needed between city and postcode
    postcode={518118}, 
    state={Guangdong},
    country={China}}
\author[3]{\textcolor{black}{Qi Zhao}}
\credit{Data Collection}

\affiliation[3]{organization={Shenzhen Institute of Advanced Technology, Chinese Academy of Sciences},
    % addressline={Radarweg 29}, 
    city={Shenzhen},
    % citysep={}, % Uncomment if no comma needed between city and postcode
    postcode={518055}, 
    state={Guangdong},
    country={China}}

\author[3]{\textcolor{black}{Dong Luo}}
\credit{Data Collection}
\author[3]{\textcolor{black}{Shuaibao Chen}}
\credit{Data Collection}

\author[3]{\textcolor{black}{Wei Chen}}
\credit{Review}
\ead{chenwei@siat.ac.cn}
\cormark[1]

\author[2]{\textcolor{black}{Xiaojiang Peng}}
\credit{Methodology, Review}
\ead{pengxiaojiang@sztu.edu.cn}
\cormark[1]

\cortext[cor1]{Corresponding authors}
% \cortext[cor2]{Principal corresponding author}

\fntext[fn1]{Equal contributions.}
% \fntext[fn2]{Another author footnote, this is a very long footnote and
%   it should be a really long footnote. But this footnote is not yet
%   sufficiently long enough to make two lines of footnote text.}

% \nonumnote{This note has no numbers. In this work we demonstrate $a_b$
%   the formation Y\_1 of a new type of polariton on the interface
%   between a cuprous oxide slab and a polystyrene micro-sphere placed

\begin{abstract}
The widespread use of various chemical gases in industrial processes necessitates effective measures to prevent their leakage during transportation and storage, given their high toxicity. Thermal infrared-based computer vision detection techniques provide a straightforward approach to identify gas leakage areas. However, the development of high-quality algorithms has been challenging due to the low texture in thermal images and the lack of open-source datasets. In this paper, we present the \textbf{R}GB-\textbf{T}hermal \textbf{C}ross \textbf{A}ttention \textbf{Net}work (RT-CAN), which employs an RGB-assisted two-stream network architecture to integrate texture information from RGB images and gas area information from thermal images. Additionally, to facilitate the research of invisible gas detection, we introduce Gas-DB, an extensive open-source gas detection database including about 1.3K well-annotated RGB-thermal images with eight variant collection scenes. Experimental results demonstrate that our method successfully leverages the advantages of both modalities, achieving state-of-the-art (SOTA) performance among RGB-thermal methods, surpassing single-stream SOTA models in terms of accuracy, Intersection of Union (IoU), and F2 metrics by 4.86\%, 5.65\%, and 4.88\%, respectively. The code and data can be found at \url{https://github.com/logic112358/RT-CAN}.
\end{abstract}

% \includegraphics{figs/grabs.pdf}

% \begin{highlights}
% \item Research highlights item 1
% \item Research highlights item 2
% \item Research highlights item 3
% \end{highlights}

% Keywords

\maketitle

\section{Introduction}

With the onset of industrialization, the use of diverse chemical gases has become widespread in industrial processes, some of which are highly toxic and pose significant hazards. Efficiently handling the transportation and storage of these gases is crucial, especially in preventing potential leakages. However, detecting gas leaks during transportation and industrial reactions presents formidable challenges, given that a considerable portion of these gases is invisible to the human eye, rendering traditional computer vision techniques ineffective.

The detection of invisible gases in industrial processes has become a focal point, driving the development of various gas detection techniques. These methods can be broadly classified into point, line, and area gas detection techniques based on their detection scope (\cite{9481237}). Point-based detection relies on gas sensors, offering advantages such as cost-effectiveness, high sensitivity, and the ability to detect gas leaks at ppm levels (\cite{morrison1987mechanism}). However, these methods are limited to a specific point, presenting challenges in identifying the source of leaks across a broader area.

\begin{figure}[t] 
\centering 
\includegraphics[width=0.47\textwidth]{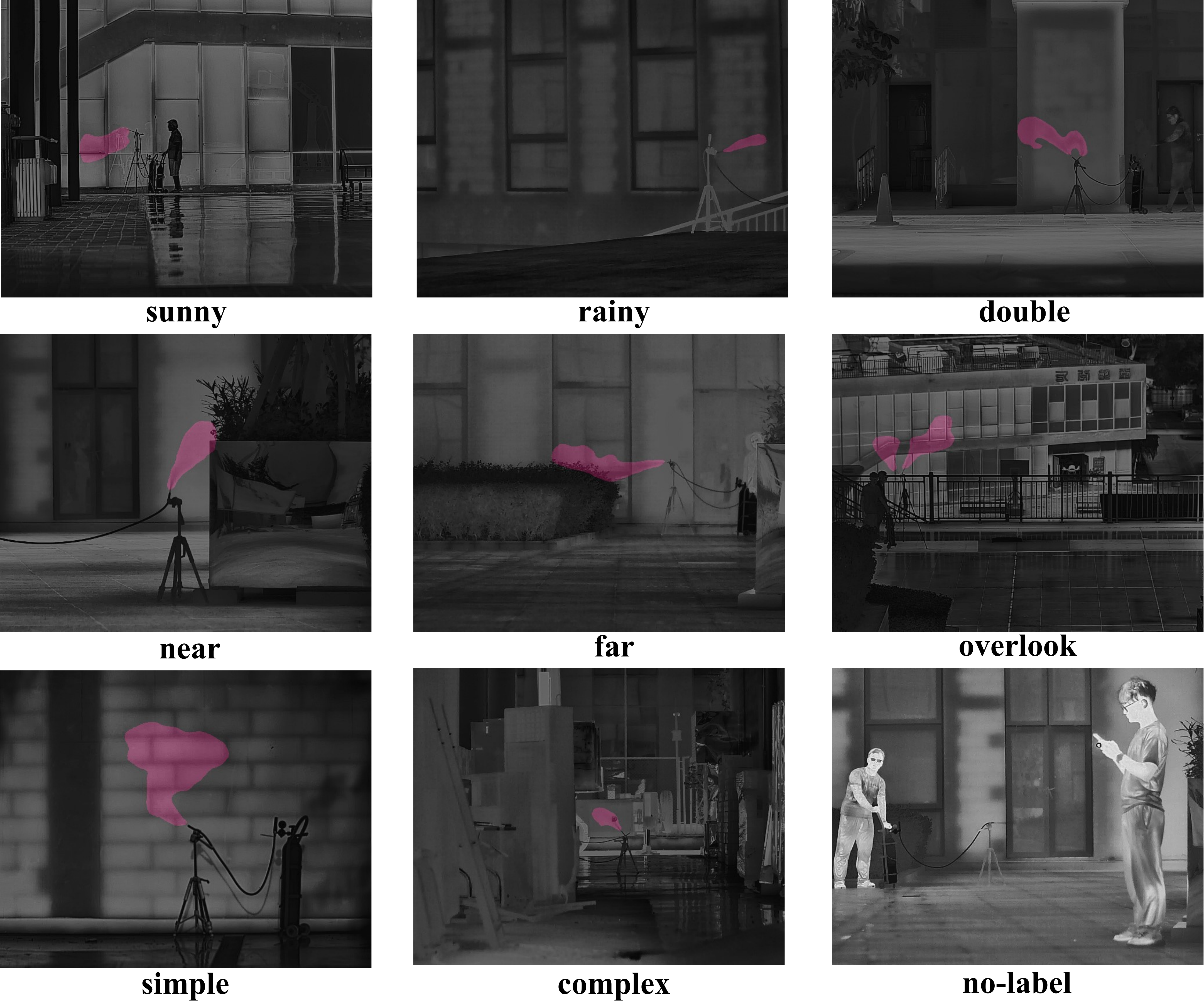}
\caption{This figure shows an overview of our Gas-DB, containing 8 kinds of scenery, containing sunny, rainy, double leakage, nearly leakage, further leakage, overlook, simple background, and complex background. The last one is the original gas image without manually annotating.} \label{database-overview} 
\end{figure}

In contrast, line-based detection techniques primarily utilize gas infrared detection through spectrometry, providing a wider detection range, clear imaging, and a better understanding of the spatial distribution of gases (\cite{gat1997spectral}). Meanwhile, area-based detection techniques aim to identify gas regions within images or video frames, representing a straightforward approach using only a camera. Nevertheless, this method's limitation lies in the fact that most gases are invisible to the human eye, making them undetectable by conventional RGB cameras. Specialized thermal infrared-based cameras are required for their visualization. However, a significant challenge arises as images captured by thermal infrared-based cameras often exhibit lower textures compared to those obtained from RGB cameras. Consequently, distinguishing between a gas area and irrelevant background noise, such as a mere black block, becomes challenging due to the reduced clarity and detail in thermal images.

In the realm of gas detection in computer vision, existing area-based detection algorithms have been explored. Traditional methodologies, as highlighted in studies by \cite{nie2021infrared, zhao2022effective} often rely on motion target and foreground segmentation techniques. While effective in certain contexts, these approaches may falter when confronted with a complex background or the introduction of motion with non-gas objects, rendering them less adept at handling intricate real-world scenarios. This limitation compromises their practical utility in diverse applications. On the other hand, deep learning-based methods have shown promise in this domain (\cite{wang2020machine, wang2022videogasnet, peng2019gas}). Nonetheless, they face challenges related to the limited diversity of available datasets, where scenes tend to be relatively uniform. This data scarcity hampers the development of deep learning models and hinders researchers from effectively evaluating the efficacy of their proposed methods.

Furthermore, another critical impediment to the progress of gas detection research is the absence of publicly available high-quality gas datasets. The lack of such datasets not only hinders researchers from adequately evaluating their methods but also restricts their ability to train and refine deep learning models on representative gas-related data. Addressing this data scarcity and promoting the availability of comprehensive gas datasets will undoubtedly foster advancements in the field and facilitate the development of more robust and accurate gas detection techniques.

In this paper, our focus centers on utilizing computer vision techniques to detect gases and addressing challenges arising from the lack of texture. It is essential to note that gases are only visible in thermal images, which, unfortunately, often exhibit significant background noise due to the underlying infrared imaging methodology. When such noise, including shadows and dark patches, emerges, even the human eye struggles to discern whether the imaging represents the gas or irrelevant background elements. However, RGB images boast rich semantic information encompassing background details, contours, shadows, and lines, making them valuable complements to thermal images.

Motivated by this intuition, we introduced a novel RGB-assisted two-stream RGB-Thermal gas detection network named \textbf{R}GB-\textbf{T}hermal \textbf{C}ross \textbf{A}ttention \textbf{Net}work (RT-CAN) to accurately detect gases. The main purpose of RT-CAN is to enhance the thermal stream's information using the RGB stream so that it could better learn the gases only shown in the thermal stream. We also propose an open source gas database, Gas-DB, consisting of 1293 pairs of spatially aligned RGB-Thermal images in multi-background and high-quality annotations (Figure \ref{database-overview}).

\section{Related Work}

In this section, we give a brief introduction to RGB-Thermal image segmentation and Thermal-based gas detection.

\subsection{RGB-Thermal Semantic Segmentation}
Algorithms for RGB-Thermal semantic segmentation can be broadly categorized into three groups:

\textbf{Early Fusion:} This method entails concatenating RGB and thermal images into 4-channel inputs, directly feeding them into a neural network for segmentation (\cite{ha2017mfnet}). However, the challenge lies in achieving effective fusion, making it difficult for the model to learn informative features from both modalities.

\textbf{Late Fusion:} In this approach, two distinct networks produce separate results, and fusion takes place after the entire network processes the inputs (\cite{liang2023explicit}). Consequently, when fusion occurs, the obtained features are highly abstract. This method is prone to losing valuable information from the inherently low-texture thermal images, exacerbated by the depth of the neural network.

\textbf{Feature Fusion:} This method employs two separate networks, integrating fusion within each network during the training process (\cite{zhou2022multispectral, deng2021feanet}). While this approach effectively considers the fusion of modal information, its effectiveness relies on the premise that the content in both thermal and RGB images is identical. Given that gases are exclusively visible in infrared images, this strategy may not be well-suited for gas detection tasks.

\subsection{Thermal-based Gas Detection}
Algorithms for thermal-based gas detection can be broadly classified into two categories: 

\textbf{Traditional Foreground Detection Methods:} These methods rely on gas infrared imaging (\cite{nie2021infrared, zhao2022effective}). They operate under simplified conditions, requiring a background image without objects and foreground images containing only moving objects. While convenient for experimentation, these conditions are overly idealized and fail to accurately capture the complexities of real-world scenarios.

\textbf{Deep Learning Based Methods:} These approaches involve employing deep learning for gas detection, encompassing gas classification, and semantic segmentation. However, datasets used in gas detection methods based on deep learning, as seen in \cite{wang2020machine} and \cite{wang2022videogasnet}, are often singular and simplistic, making them susceptible to overfitting, hindering the generalization of these methods to real-world scenarios.

\begin{figure}[t]  \centering \includegraphics[width=0.46\textwidth]{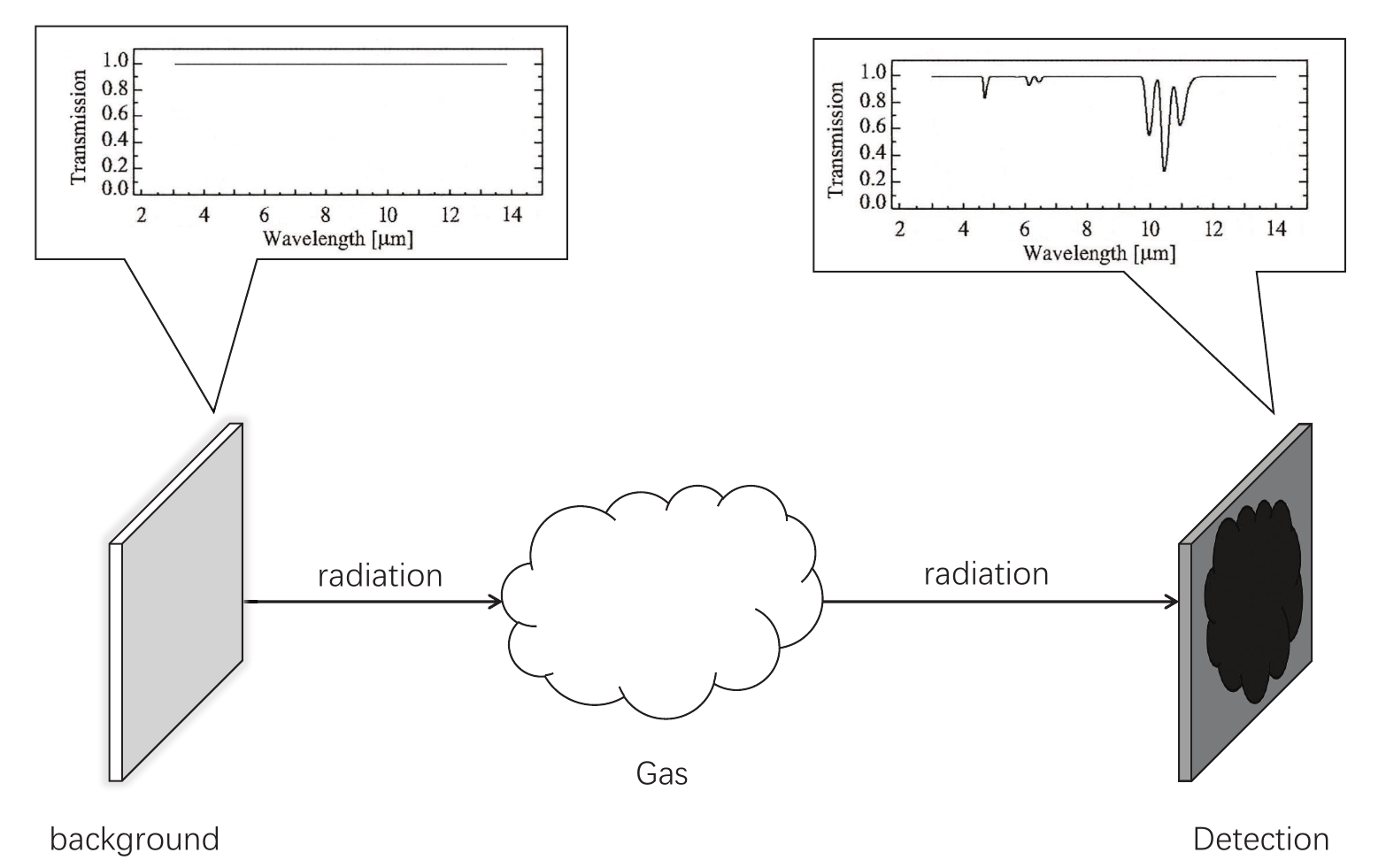} \caption{This figure illustrates the imaging mechanism of gas in thermal infrared images, where gas is imaged through the absorption of specific wavelengths of infrared radiation.} \label{imaging-theory} \end{figure}

\section{Gas-DB: A Comprehensive Gas Detection Dataset}
The only open source dataset available in the research community, GasVid, offers thermal videos of gas releases at five distances ranging from 4.6 meters to 18.6 meters (\cite{wang2022videogasnet}). However, it presents an idealized representation with the simplest background of the sky that does not fully encompass the complexities of real-world environments. This limitation poses challenges in developing robust and generalizable gas detection algorithms. To overcome this shortfall, there is a critical need for a high-quality dataset that faithfully replicates real-world scenarios, encompasses diverse gas-related scenes, and features accurate and detailed annotations.

\begin{table}[h]
  \caption{The comparsion of properties between existing gas detection datasets and the Gas-DB.}
  \label{datasets}
    \begin{tabular}{cccc}
    \hline
     & Database  & GasVid  & Gas-DB(Ours)  \\
     \hline
     & Modality & Thermal & RGB-Thermal \\
     & Scenery  & 1 & 8 \\
     & Background & Sky & Real-world scenery  \\
     & Task & Classification & Segmentation  \\
     & Distance & 4.6-18.6m & 3-20m  \\
     & Quantity (video) & 31 & 19  \\
     & Availability & Public & Public  \\
     \hline
     
    \end{tabular}
\end{table}

\begin{figure}[t] \includegraphics[width=0.46\textwidth]{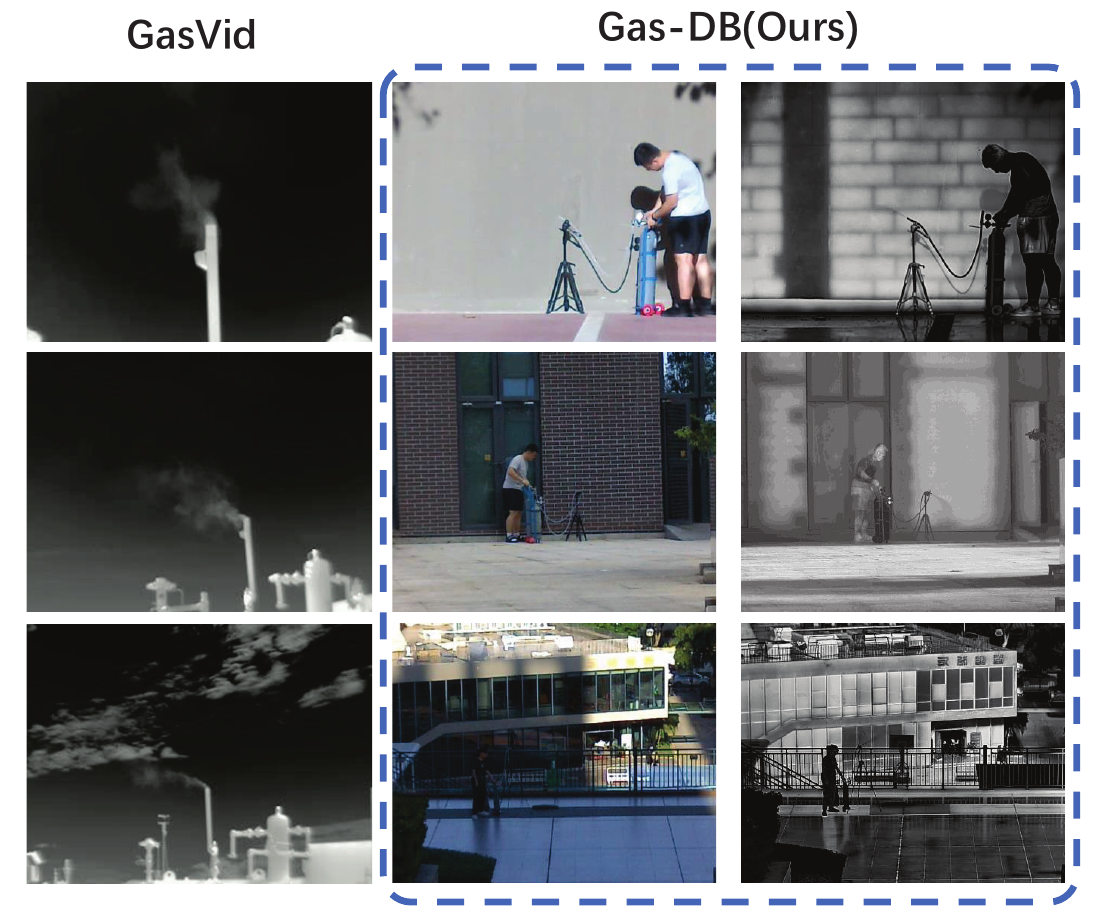} \caption{The comparision of the scenery of existing database and the Gas-DB.} \label{Gas-DB} \end{figure}

In response to this imperative, we introduce Gas-DB, a meticulously constructed gas dataset tailored explicitly for gas detection tasks. Gas-DB is the outcome of thorough data curation and annotation processes, ensuring its reliability and efficacy in evaluating gas detection algorithms. Comparison of between Gas-DB and GasVidis illustrated

\subsection{Dataset Collection and Annotations}
The foundational principle of gas infrared imaging is intricately tied to the absorption spectrum of gas molecules. When a heat source emits radiation through a gas, the gas molecules or atoms selectively absorb radiation that aligns with their specific energy gaps, allowing only certain radiations to pass through without interaction. Analyzing the attenuation of transmitted radiation reveals pronounced spectral selectivity in the gas under observation, leading to noticeable radiation contrast between gas and non-gas regions. In the imaging results of an infrared detector, this contrast manifests as distinguishable differences in pixel grayscale values. These variations enable the detection and visualization of gas areas in captured infrared images. By leveraging the absorption characteristics of gas molecules—chemical gases, in particular, exhibit absorption characteristics in the far-infrared wavelength range of 3-14 micrometers \cite{9481237, gordon2022hitran2020}—infrared imaging emerges as a powerful tool for detecting and analyzing invisible gases in various industrial and environmental settings (Figure \ref{imaging-theory}). 

To construct Gas-DB, we initiate the process by capturing videos from real-world environments using a thermal infrared camera (3-14 micrometers) and an RGB camera, both with a resolution of 640x512. To ensure accurate alignment, the optical axes of the devices are secured, and pixel-level registration is performed between the RGB and infrared images. To simulate diverse practical scenarios, we collect footage encompassing eight different types of gas leakage scenes. These scenes include various environmental conditions such as clear and overcast weather, scenarios with multiple simultaneous leakage points (double), top-down views of leakage points (overlook), as well as scenes with both simple and complex backgrounds (Figure \ref{database-overview}). In total, we collect over 19 videos depicting these scenarios.

Subsequently, the boundaries of gas regions may appear blurred due to varying background thermal radiation levels in different environments and the inherent lack of texture in infrared images. To address this challenge, we employ a Visibility Restoration Algorithm enhancement technique (\cite{5459251}) alongside the \textit{LabelMe} tool for the annotation process. In cases where the diffusion range of gas cannot be directly determined by the human eye, we utilize background subtraction to precisely annotate gas regions. Following meticulous collection and annotation, the Gas-DB dataset comprises a total of 1293 images.

\begin{figure*}[t] \centering \includegraphics[width=1\textwidth]{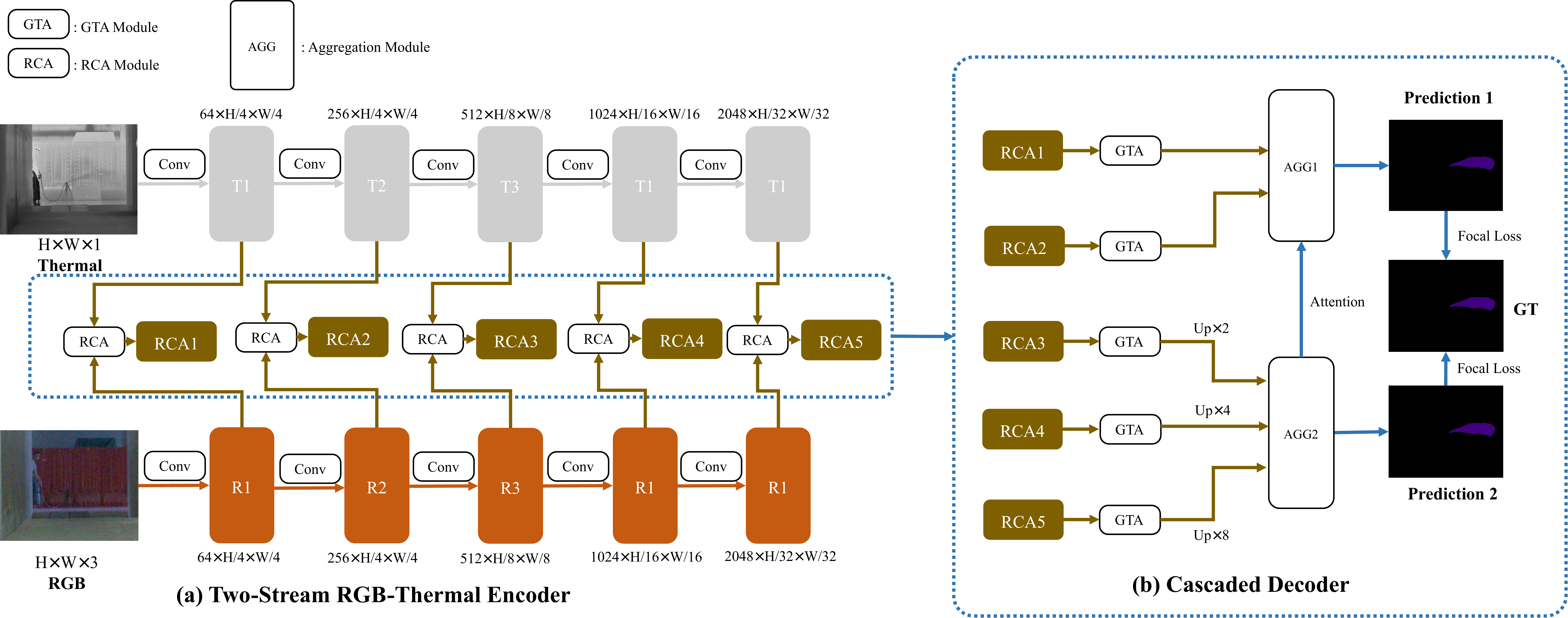} \caption{Illustration the architecture of RGB-Thermal Two Stream Cross Attention Network. (a) Two stream RGB-ThermaR Cl Encoder, (b) Cascaded Decoder.} \label{model} \end{figure*}

\section{RT-CAN: Enhancing Gas Detection}
We present the \textbf{R}GB-\textbf{T}hermal \textbf{C}ross \textbf{A}ttention \textbf{Net}work (RT-CAN) designed specifically for the automatic detection of vision-invisible gases. Existing gas detection methods, including those employing deep learning, encounter challenges in effectively addressing gas detection from thermal images due to factors such as the lack of texture. These challenges often lead to misclassifications, where features like black blocks or shadows are erroneously identified as gas regions. Furthermore, prevailing RGB-Thermal object recognition models predominantly focus on fusing RGB and Thermal information, overlooking the potential effectiveness of leveraging a single camera type when information is sufficiently rich. Conversely, in scenarios with limited information, simple fusion techniques may fall short of producing satisfactory results.

To overcome these limitations, we propose a novel two-stream RGB-Thermal cross attention gas detection network with a \textbf{R}GB-assisted \textbf{C}ross \textbf{A}ttention (RCA) Module, specifically designed to address the challenges in RGB-Thermal recognition.

For more precise segmentation, we employ a cascade decoder architecture in conjunction with the \textbf{G}lobal \textbf{T}extural \textbf{A}ttention (GTA) module for decoding. This strategic combination enables meticulous pixel-level segmentation of gas regions, resulting in significantly enhanced detection performance. By capitalizing on the synergy of information between RGB and Thermal streams, our proposed RT-CAN is poised to overcome the limitations faced by traditional methods and provides a more effective solution for detecting vision-invisible gases.

\begin{figure}[t] \centering \includegraphics[width=0.48\textwidth]{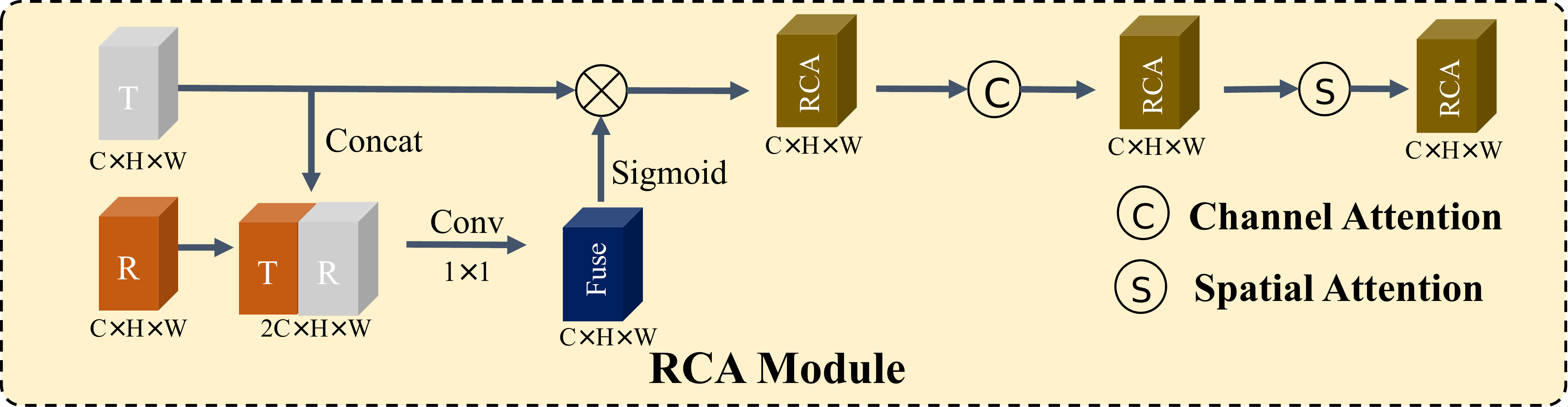} \caption{Proposed RGB-assisted Cross Attention (RCA) Module.} \label{RCA} \end{figure}

\subsection{Architecture}
We adopt a classical encoder-decoder network architecture for gas detection, as illustrated in Figure \ref{model}. The proposed RT-CAN features two symmetric input encoding streams for effective feature extraction. Each encoding block's output is fed into a \textbf{R}GB-assisted \textbf{C}ross \textbf{A}ttention (RCA) module, facilitating the fusion of information from RGB to thermal features.

The two encoding streams are based on ResNet-50 (\cite{He_2016_CVPR}) as the backbone, with modifications to the input channels to accommodate RGB and thermal images. The outputs from the five RCA modules are then directed to a cascaded decoder, providing multi-scale representations of gas regions. Subsequently, the Global Textural Attention (GTA) module is employed to capture global texture information. The final detection is performed using two Aggregation heads (\cite{liang2023explicit}), acting as the ultimate detectors.

To enhance the model's ability to identify gas regions accurately, we introduce an attention mechanism between the Aggregation head with deeper high-level features and the head with low-level features. The final prediction is obtained by averaging the outputs of the two detector heads, ensuring a comprehensive and refined gas detection outcome.

\subsection{RGB-assisted Cross Attention (RCA) Module}

The main challenge in previous RGB-Thermal approaches lies in their simplistic fusion of RGB and thermal modalities for gas detection. Since gases are exclusively visible in thermal infrared images, traditional fusion strategies prove ineffective.

To address this issue, we introduce the RGB-assisted Cross Attention (RCA) module, depicted in Figure \ref{RCA}. The module begins by concatenating thermal and RGB features into a fused feature. Subsequently, a 1 $\times$ 1-kernel convolution is employed to extract relevant information within the fused feature. The resulting feature undergoes rescaling through a sigmoid function, bringing its values within the range of 0-1. Finally, this rescaled feature is multiplied back with the original thermal feature. The entire process can be expressed as follows:
$$
RCA = T \otimes Sigmoid(Conv_{1 \times 1}(concat(T, R)))
$$
where $RCA$ is the output of RCA module, $T$, $R$ is the thermal feature and RGB feature respectively,  $\otimes$ denotes element-wise multiplication.

Then, we use channel attention and spatial attention mechanism sequentially:
$$
RCA = f_c(f_s(RCA))
$$
where $f_c$ and $f_s$ denote the channel attention and spatial attention mechanism respectively.

\subsection{Cascaded Decoder}
Recognizing the limited textual features in thermal infrared images and the information loss incurred during the encoder's downsampling process, we implement a cascaded decoder to harness multi-level information. This approach involves integrating both shallow and deep-level encoder information, amplifying the discernment of crucial features and enhancing the model's capacity for pixel-level classification.

Following the enhancement of features by the GTA module, two cascaded AGG (\cite{liang2023explicit}) modules are employed as the final detection head. As illustrated in Figure \ref{model}, the AGG module from the deep-level features operates as an attention block to the AGG module from the shallow-level features. Ultimately, by combining the outputs of the two AGG modules, we generate the final prediction.

\subsection{Global Textual Attention (GTA) Module}

\begin{figure}[t] \centering \includegraphics[width=0.46\textwidth]{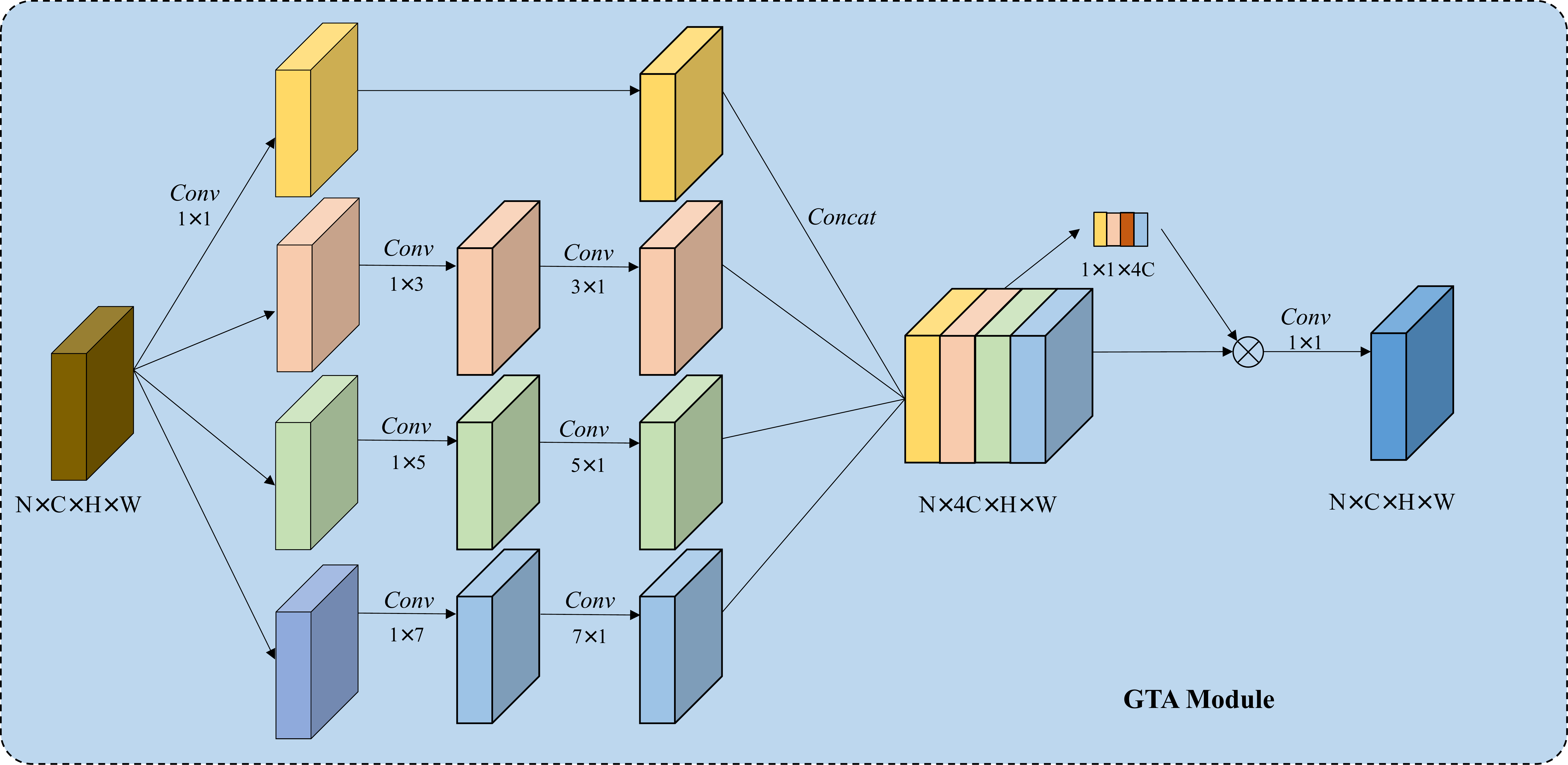} \caption{Proposed Global Textual Attention (GTA) Module.} \label{GTA} \end{figure}

In the decoder, we introduce the Global Textual Attention (GTA) module, as depicted in Figure \ref{GTA}. This module utilizes various convolution operations with multi-level kernels and concatenates the resulting multi-level features. Subsequently, a self-channel attention mechanism is applied. This process allows the model to extract multi-scale features, enhancing its ability to capture diverse contextual information.

\subsection{Implementations}
Our methods were implemented in PyTorch \cite{Paszke_PyTorch_An_Imperative_2019}. The model was initialized using ResNet (\cite{He_2016_CVPR}), pre-trained on the ImageNet dataset (\cite{deng2009imagenet}). The loss function is a combination of the Dice loss (\cite{milletari2016v}) term and a Soft Cross Entropy loss (\cite{yi2019probabilistic}), both weighted by a scalar factor of 0.5, formulated as:
$$l = 0.5 \times l_{Dice} + 0.5 \times l_{SCE}$$
where $l$ denotes the loss function, $l_{Dice}$, $l_{SCE}$ denotes Dice loss and Soft Cross Entropy respectively. Regarding the training of the model on the Gas-DB dataset for 50 epochs, and the best-performing model on the validation set is selected for evaluation.

During training and testing, a random partition of 80\% of the images is designated as the training and validation set, while the remaining 20\% forms the test set. The initial learning rate is initialized at 0.02, with Momentum and weight decay values set to 0.9 and 0.0005, respectively. A batch size of 4 is employed, and the chosen optimizer is stochastic gradient descent (SGD) (\cite{montavon2012neural}), complemented by ExponentialLR for learning rate decay. All training and testing procedures are conducted using the PyTorch framework (\cite{paszke2019pytorch}) on an NVIDIA Tesla V100S GPU.

\subsection{Metrics}
In the context of the gas detection task, we utilize conventional metrics such as Accuracy (Acc) and Intersection over Union (IoU) to quantify the accuracy and precision of segmentation, respectively. These metrics are calculated as follows:
$$
Acc = \frac{TP + TN}{TP + TN + FP + FN}
$$
and IoU is calculated as:
$$
IoU = \frac{TP}{TP + FP + FN}
$$
where $TP$, $TN$, $FP$, and $FN$ denote True Positives, True Negatives, False Positives, and False Negatives respectively.

In gas detection tasks, the F2 score is introduced as a crucial metric due to the unique risks associated with undetected gas leaks. The F2 score combines precision and recall, emphasizing the need to minimize false negatives (missed gas detections). This is particularly relevant in gas detection, where ensuring a high recall rate is essential for promptly identifying potential leaks, mitigating environmental and health risks. While accuracy is a common metric, the F2 score provides a more nuanced evaluation, aligning with the specific challenges and priorities of gas detection tasks. It is calculated as follows:
$$
F2 = \frac{(1 + \beta^2) \cdot Precision \cdot Recall}{\beta^2 \cdot Precision + Recall}
$$
where $\beta $ is a parameter that controls the relative importance of precision and recall. A higher value of $\beta$ emphasizes recall more, in our experiment, we set $\beta$ to 2.

\begin{table}[h]
  \caption{Comparisons of different segmentation models, '4c' denotes using 4 channels (RGB and thermal) as input, and '1c' denotes using 1 channel (thermal) as input.}
  \label{Results}
  \scalebox{0.85}{
  \begin{tabular}{cccc}
    \toprule
    \multirow{2}{*}{Methods} & \multicolumn{3}{c}{Metrics} \\
    & $ Acc $ ($\uparrow$) &	$ IoU $ ($\uparrow$) &	$ F2 $ ($\uparrow$)\\
    \midrule
    PSPNet (4c)                  &   {46.0164}  &   {34.3776}   &   {47.9465} \\
    Segformer (4c)              &   {56.6855}  &   {37.0581}   &   {55.6122} \\
    YOLOv5 (4c)                &   {71.4862}  &   {48.8109}   &   {69.0099}\\
    \toprule
    \multirow{1}{*}{RGB-Thermal} & \multicolumn{3}{c}{Metrics} \\
    \midrule
    MFNet                   &   {59.3935}  &   {37.1752}   &   {57.2015} \\
    FEANet                &   {63.1716}  &   {50.8582}   &   {64.8070} \\
    RTFNet                &   {67.5185}  &   {50.9560}   &   {67.5155} \\
    EAEFNet              &   {74.5951}  &   {53.4445}   &   {72.5394}\\
    \textbf{Ours(ResNet152)}        &   {74.7540}  & \textbf{56.5167}   & {73.7186} \\
    \textbf{Ours(ResNet50)}         &   \textbf{76.3404}  & {54.4604}   & \textbf{73.8993} \\
    
    \bottomrule
  \end{tabular}
  }
\end{table}

\subsection{Quantitative Analysis}
In our quantitative analysis, we compare our proposed method with existing approaches, encompassing RGB one-stream segmentation models like PSPNet (\cite{zhao2017pyramid}), SegFormer (\cite{xie2021segformer}), YOLOv5 (\cite{glenn_jocher_2020_4154370}), and RGB-Thermal two-stream segmentation models, including MFNet (\cite{ha2017mfnet}), RTFNet (\cite{sun2019rtfnet}), and EAEFNet (\cite{liang2023explicit}).

To ensure a fair comparison, we implement RT-CAN using ResNet-50 and ResNet-152 as backbones. The results are presented in Table \ref{Results}.

The findings highlight the subpar performance of traditional one-stream models, regardless of whether a single thermal image or RGB images stacked into a 4-channel input are employed. These approaches exhibit significant limitations, showcasing a notable performance gap when contrasted with RGB-Thermal two-stream models. Additionally, our proposed model outperforms the RGB-Thermal fusion-focused model significantly.

It is noteworthy that EAEFNet (\cite{liang2023explicit}), despite displaying a performance discrepancy compared to our model, exhibits superior performance over other models. This difference in performance could be attributed to EAEFNet's emphasis on leveraging reciprocal information exchange between RGB and Thermal modalities. This approach shares similarities with our methodology, contributing to its enhanced performance. Moreover, from the experimental data, when the model backbone transitions from ResNet-50 to ResNet-152, the corresponding metrics do not exhibit improvement; instead, a certain level of degradation is observed. This phenomenon is likely attributed to the deepening of the model, which amplifies the low-texture characteristics in infrared images, leading to the loss of gas information and consequently impacting the model's performance.

\begin{figure*}[t] \centering \includegraphics[width=1\textwidth]{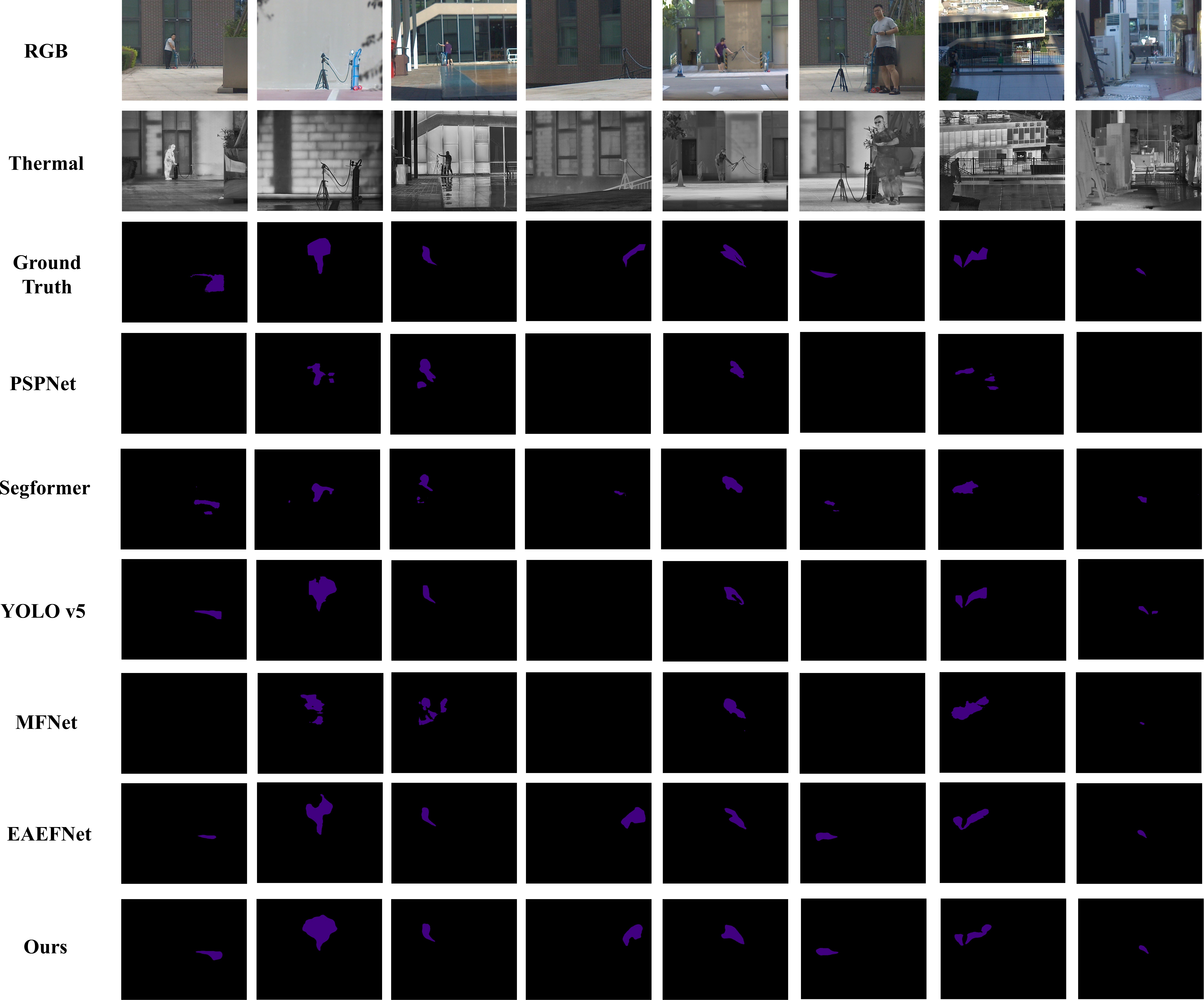} \caption{The visualization of the prediction comparisons from different methods, according to the rows from top to bottom in order: RGB; Thermal; Ground Truth; PSPNet; Segformer; YOLOv5; MFNet; EAEFNet; Ours.} \label{result} \end{figure*}

\subsection{Qualitative analysis}
Figure \ref{result} visually presents the predictions generated by various methods, offering insights into their performance. In comparison to one-stream methods, our proposed RT-CAN demonstrates a significant improvement. One-stream methods exhibit frequent instances of both missed detections and false positives. These occurrences can be attributed to inherent limitations in thermal imaging, causing the model to erroneously identify shadows and dark regions as gas regions. RT-CAN, with its integration of rich information from RGB images, effectively distinguishes between such noise and actual gas regions, resulting in more accurate detections.

In contrast to two-stream methods, they exhibit issues such as missed detections and imprecise detection. This is attributed to the assumption underlying these methods that information in the two images is somewhat consistent, leading to simplistic fusion techniques. However, gas occurrences are exclusive to thermal images. Therefore, straightforward fusion introduces noise. RT-CAN, incorporating RGB information through the RCA module, skillfully supplements thermal information, circumventing mutual interference between the modalities. This approach ensures a more accurate and robust detection process, setting RT-CAN apart from other methods.

\subsection{Ablation Study}
In this subsection, we experimentally analyze the effectiveness of our Gas-DB dataset and the proposed RT-CAN.

\begin{figure}[t] \centering \includegraphics[width=0.46\textwidth]{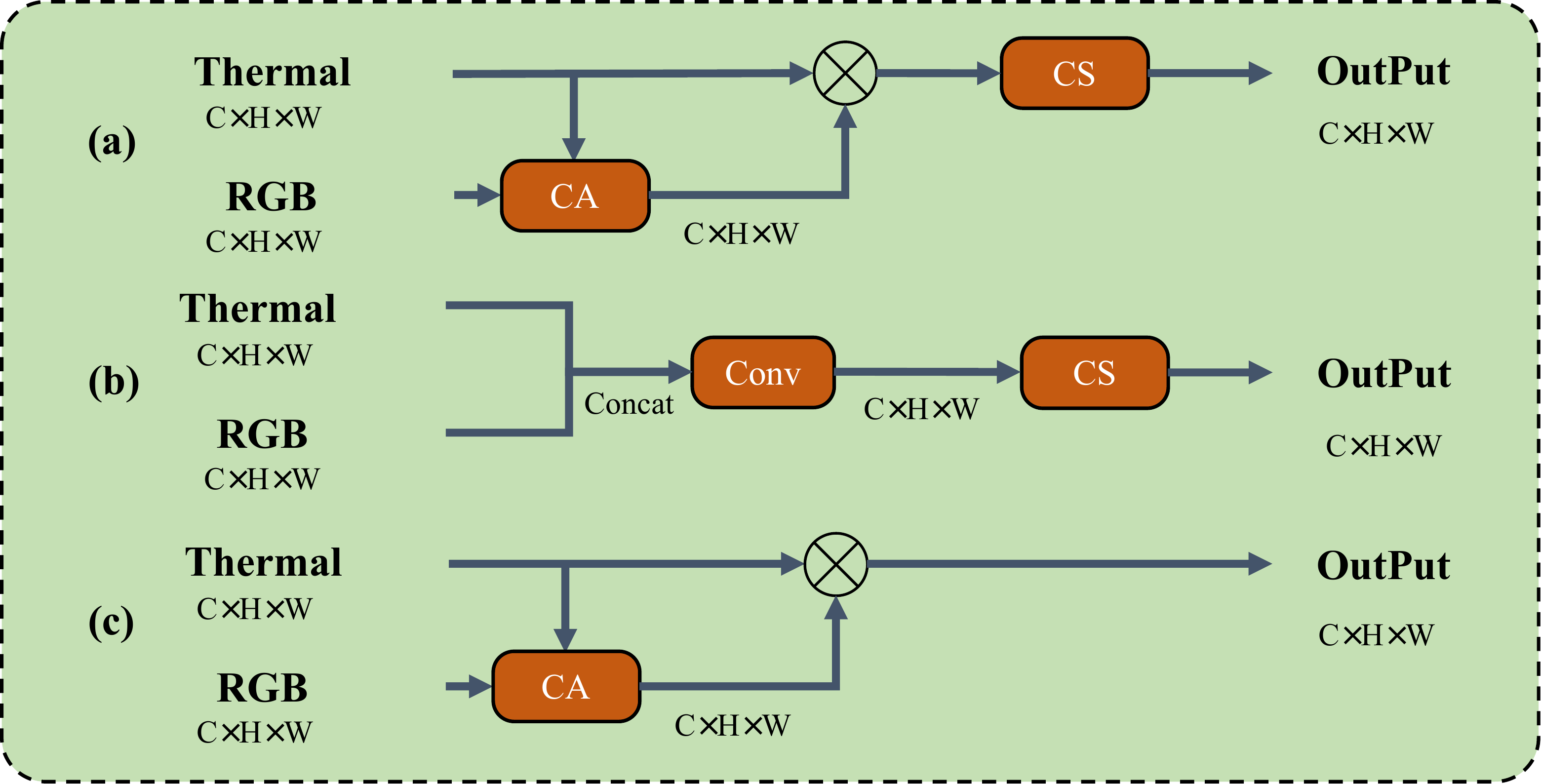} \caption{Three different schemes of the 
ablation study of RCA module. "CA" stands for Cross Attention module, and "CS" stands for the combination of channel attention and spatial attention.} \label{abs} \end{figure}

\textbf{Comparison of Modified Modules:} The most significant difference between the proposed RT-CAN and the previous baseline framework is that we make the RGB-Thermal feature fusion as RGB-Assisted cross attention to assist the change of the thermal feature rather than direct fusion or concatenation. As shown in Figure \ref{abs}, we implemented three ablation experiments on different schemes of the RCA module:

\begin{itemize}
  \item \textbf{Scheme A:} The proposed architecture, where RGB and Thermal features undergo a Cross Attention module, followed by multiplication with the original Thermal features, and then further processed through a combination of channel attention and spatial attention.
  
  \item \textbf{Scheme B:} We eliminate the RCA module and concatenate the thermal and RGB features for subsequent convolution instead.
  
  \item \textbf{Scheme C:} We exclude the channel attention and spatial attention modules.
\end{itemize}

The results of the ablation experiments are shown in Table \ref{Result2}. All the above results demonstrate that our method is effective and powerful, indicating our method can be used as a trustful gas detection technique.

\begin{table}[h]
  \caption{Results of ablation studies. 'A', 'B' and 'C' stands for different setting scheme.}
  \label{Result2}
    \begin{tabular}{ccccc}
    \hline
     & Scheme  & ACC  & IoU & F2  \\
     \hline
     & A(Ours) & \textbf{76.3404} & \textbf{54.4604} & \textbf{73.8993} \\
     & B & 72.5169 & 52.8349 & 71.1272 \\
     & C & 73.3471 & 52.1753 & 71.3597 \\
     \hline
     
    \end{tabular}
\end{table}

\section{Conclusion}
In this paper, we introduced the RGB-Thermal Cross Attention Network (RT-CAN) for vision-invisible gas detection, comprising both RGB and thermal streams. Diverging from traditional methods that employ a simple fusion strategy, our approach incorporates the RGB modality to assist the thermal modality in feature fusion through the RCA module, coupled with the GTA module for global textual feature extraction. Experimental results showcase that our method achieves state-of-the-art (SOTA) performance in gas detection. Ablation studies on the design choices further validate the effectiveness of the proposed model. Additionally, we introduced the first well-annotated real-world open-source RGB-Thermal gas database, Gas-DB, anticipated to significantly benefit the community's future research in real-world gas detection applications.


\bibliographystyle{model1-num-names}
% \bibliographystyle{cas-model2-names} 

\bibliography{cas-refs}


\end{document}